\documentclass{article}

\usepackage[nonatbib, final]{nips_2017}

\usepackage[utf8]{inputenc} 
\usepackage[T1]{fontenc}    
\usepackage{hyperref}       
\usepackage{url}            
\usepackage{booktabs}       
\usepackage{amsfonts}       
\usepackage{nicefrac}       
\usepackage{microtype}      

\usepackage{epsfig}
\usepackage{graphicx}
\usepackage{amsmath}
\usepackage{amssymb}
\usepackage{subfig}
\usepackage{url}
\usepackage{color}

\title{Poverty Prediction with Public Landsat 7 Satellite Imagery and Machine Learning}

\author{
  Anthony Perez \\
  Department of Computer Science \\
  Stanford University \\
  Stanford, CA 94305 \\
  \texttt{aperez8@stanford.edu} \\
  \And
  Christopher Yeh \\
  Department of Computer Science \\
  Stanford University \\
  Stanford, CA 94305 \\
  \texttt{chrisyeh@stanford.edu} \\
\And
George Azzari \\
Department of Earth System Science\\
Stanford University\\
Stanford, CA - 94305\\
{\tt\small gazzari@stanford.edu}
\And
Marshall Burke \\
{Department of Earth System Science}\\
Stanford University\\
Stanford, CA - 94305\\
{\tt\small mburke@stanford.edu}
\And
David Lobell \\
{Department of Earth System Science}\\
Stanford University\\
Stanford, CA - 94305\\
{\tt\small dlobell@stanford.edu}
  \And
Stefano Ermon \\
{Department of Computer Science}\\
Stanford University\\
Stanford, CA - 94305\\
{\tt\small ermon@cs.stanford.edu}
}

\begin{document}

\maketitle






\begin{abstract}
Obtaining detailed and reliable data about local economic livelihoods in developing countries is expensive, and data are consequently scarce. Previous work has shown that it is possible to measure local-level economic livelihoods using high-resolution satellite imagery. However, such imagery is relatively expensive to acquire, often not updated frequently, and is mainly available for recent years. We train CNN models on free and publicly available multispectral daytime satellite images of the African continent from the Landsat 7 satellite, which has collected imagery with global coverage for almost two decades. We show that despite these images' lower resolution, we can achieve accuracies that exceed previous benchmarks. 
\end{abstract}

\section{Introduction}


Policy makers and philanthropic organizations rely on data about local economic livelihood to direct their efforts in places that most need aid \cite{un15}, \cite{devarajan13}, \cite{jerven13}. Traditionally, such data has come from expensive and logistically challenging household surveys. This has meant that nationally-representative surveys are conducted only intermittently, with 39 of 59 African countries conducting fewer than two surveys during the years 2000 to 2010 from which nationally representative poverty measures could be constructed \cite{un15}. As a result, both policymakers and researchers lack key data with which to target anti-poverty programs or to measure their effectiveness. 





Previous work \cite{Jean16} introduced transfer learning methods for estimating economic livelihood indicators in 5 African countries from satellite imagery: Malawi, Nigeria, Rwanda, Tanzania, and Uganda. Reasoning that nighttime light intensity is correlated with urban developments, Jean et al. trained a convolutional neural network (CNN) to predict nighttime light intensity from daytime satellite images. They then trained simpler models on image features extracted by the CNN to estimate an Asset Wealth Index (AWI). \cite{Jean16} found that the CNN features were useful in predicting asset wealth within the poorest segment of the population, especially when compared to established methods.


Building on \cite{Jean16}, we introduce the following contributions:
\begin{enumerate}
\item \textbf{Publicly available, freely distributable satellite imagery with a long time series.} Satellite images pulled from the Google Static Maps API (as done by \cite{Jean16}) are limited by Google's licensing terms, cannot be re-distributed, and do not have information about the date when each satellite image was taken. In contrast, we use publicly available, freely distributable multi-spectral satellite imagery from Landsat 7, available from 1999 to today.

\item \textbf{Multi-spectral satellite imagery.} Although Landsat 7 images are lower resolution than the zoom-level 16 Google Static Maps images used by Jean et al. (15-30m/px instead of approx. 2.5m/px), we achieve equivalent or better results by incorporating additional spectral bands beyond the visible spectrum information available in Google Static Maps.
\end{enumerate}

\section{Data and Preprocessing} \label{section:Data}
We created yearly composite satellite images of the African continent from 2004 to 2015 captured by the Landsat 7 satellite \cite{landsat}. Each annual composite is created by taking the median of each cloud free pixel available during that year. This preprocessing has seen success in similar applications as a method to gather clear satellite imagery. 


Landsat 7 images have 9 spectral bands (we use both the low-gain and high-gain Thermal band) ranging in resolution from 60 meters per pixel (m/px) to 15m/px \cite{landsat}. We apply pan-sharpening to the RGB bands \cite{pansharpening_paper} \cite{pansharpenening_website} to produce 15m resolution versions, as others have shown this technique to be beneficial in a variety of satellite imagery tasks \cite{pansharpening_paper}.

As in \cite{Jean16}, we use transfer learning with nighttime lights labels coming from DMSP \cite{DMSP}. We bin the nighttime light intensities into 3 classes: low, medium, and high brightness. Likewise, our AWI labels come from Demographic and Health Surveys (DHS) for multiple countries in Africa for the years of interest. However, the coverage in these surveys is sparse compared to the large amounts of satellite image data that we have at our disposal.

\begin{figure}[h]
  \centering
  \includegraphics[width=0.8\textwidth]{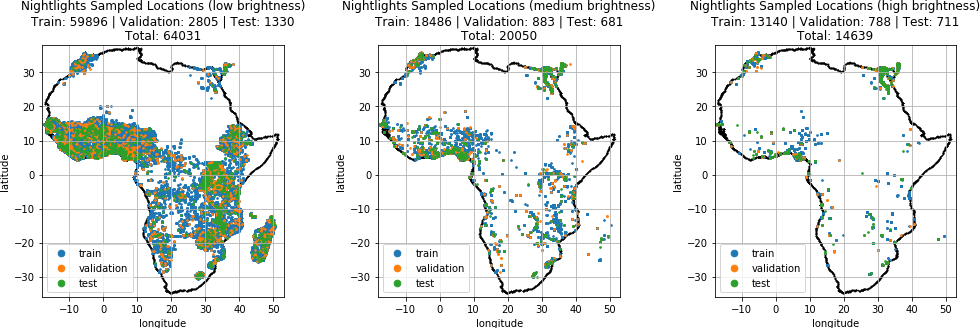}
  \caption{Sampling of daytime satellite images based on their nighttime light intensity: (left) low - class 0, (middle) medium - class 1, (right) high - class 2. The locations are divided into training (blue), validation (orange) and test (green) splits. Note this visualization shows overlap as images displayed at their actual size would be difficult to see.}
  \label{fig:nl_sampling_locations}
\end{figure}

We sample training imagery (Figure ~\ref{fig:nl_sampling_locations}) more densely near locations where labeled survey data is available in order to create more similar image distributions across the transfer learning domains. We take care dividing our sampled images into training, validation, and test splits such that there is no spatial overlap among the splits (though some images within each split may overlap).
 
\section{Methods}


\subsection{CNNs}

Most existing CNN models are designed to work with 3-channel RGB images and thus are not directly compatible with multi-band satellite images. Thus, we adapted several existing architectures to work on multi-band satellite images: 18- and 34-layer ResNets \cite{resnet_preact_paper} and VGG-F \cite{vgg_paper}. We trained each model using all bands and using only the RGB bands. When using only the RGB bands, we initialized the CNNs with weights pre-trained on the ImageNet dataset \cite{imagenet_cvpr09}. When using all 9 bands, we modified the filters of the first convolutional layer to have a depth of 9 instead of a depth of 3. In other words, the dimension of the weights becomes $[F, F, 9, 64]$ instead of the usual $[F, F, 3, 64]$, where $F$ is the filter size: 7 for ResNets, 11 for VGG-F. The weights for the RGB bands are initialized as usual, and the weights for the non-RGB bands are set to the mean of the 3 RGB weights at the same position in the same filter.

We trained each CNN model to predict the nighttime light intensity class (0, 1, or 2) from Landsat 7 daytime satellite images. We run training for 60 epochs and choose the weights from the epoch in which the model achieved the highest accuracy on the validation split. Then, we run the trained models on images corresponding to the DHS survey locations and save the features extracted by the last layer of the CNN.

\subsection{Multiple Resolution Imagery}

A challenge in dealing with satellite imagery is that the bands of the imagery may have different resolutions, as explained above. A naive workaround is to upsample all bands to the same, highest resolution, which may cause artifacts due to duplicated pixel values and poor utilization of pretrained weights.

Instead, we upsample all bands to the highest resolution of 15m/px using Nearest Neighbors and apply dilated convolutions (also called atrous convolutions) \cite{dilation_paper_original} in the first layer of the CNN. At a high level, the goal of modifying the first layer implementation is to preserve the ability to initialize the network from weights pre-trained on RGB image datasets (such as Imagenet) while removing potential artifacts caused by the mismatched resolutions of the multi-spectral imagery. The dilated convolutional layers we implement vary with with the overall model architecture.

The VGG-F model begins with an 11x11 stride-4 convolution in its first layer, whose weights are a [11, 11, 9, 64] tensor initialized from ImageNet as described above. Then the convolutional windows in the first layer are dilated to match the resolution of the original bands: the filters corresponding to the 15m bands are dilated at a rate of 1, the 30m bands at rate 2, and the 60m bands at rate 4. A stride of 4 is still applied, but no pixels are "skipped" by the convolution because nearest-neighbor upsampling to 15m results in a duplication of pixel values at a factor equivalent to the dilation rate. For example, each pixel in the 60m bands is replaced with a 4x4 block of pixels of the same value in the upsampled image. The dilation of 4 applied to this band realigns the convolutional window to the original pixel values, and thus the stride of 4 only skips the duplicated values.

The ResNet family of models, as described in \cite{resnet_preact_paper}, use a 7x7 stride-2 convolution in the first layer of the network. Our specialized implementation uses a stride of 1 and adds dilation in the same manner as our VGG-F first layer implementation.


\subsection{From Image Features to Poverty Metric}
We also tested several models for predicting poverty metrics from the image features extracted by the CNNs, including ridge regression and gradient-boosted trees (GBTs). We trained each model with leave-one-country-out cross-validation.

\section{Results}

Table~\ref{tab:corr} provides a quantitative comparison of several models trained using our methods as well as results from \cite{Jean16}. We also show results training ridge regression and GBT models on only the scalar nightlights value from each DHS survey location. The squared correlation coefficient ($r^2$) between nightlights and the AWI was 0.57, which several models we trained were able to surpass. However, applying a non-linear method, such as GBTs, to predict the AWI from nightlights yields a stronger $r^2$ value of 0.66.

\begin{figure}[h]
  \centering
  \subfloat[][]{
    \label{fig:poorest_percent}
    \includegraphics[width=0.45\linewidth]{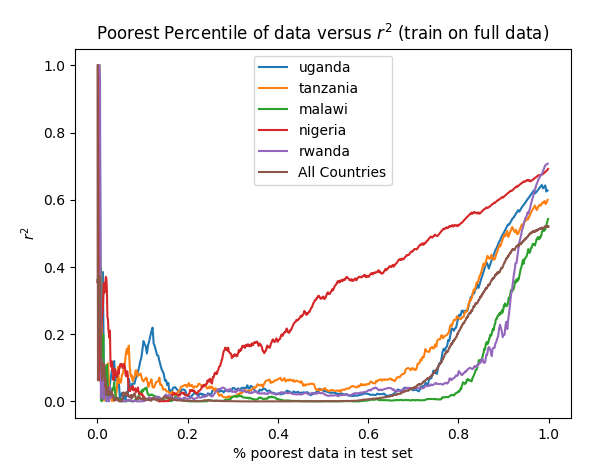}
  } \hfill
  \subfloat[][]{
    \label{fig:pool_poorest_percent}
  	\includegraphics[width=0.45\linewidth]{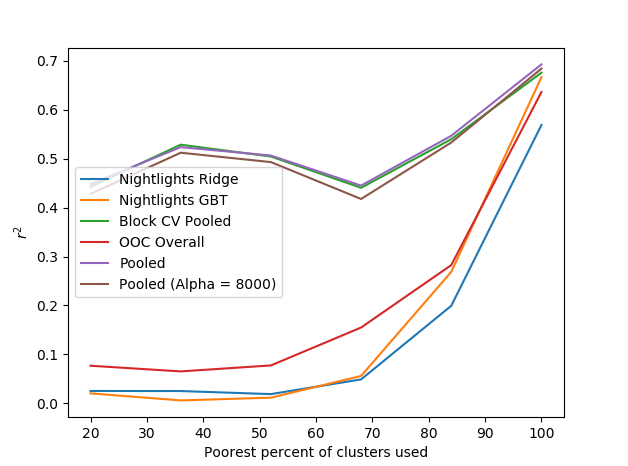}
  } \hfill
  \caption{(a) Results of leave-one-country-out training of a ridge regression model on DHS survey data for each left-out country. ``All Countries'' indicates the aggregated predictions across all 5 countries. We compute the $r^2$ value on only the datapoints below a wealth percentile threshold within the test set. The horizontal axis plots the wealth percentile threshold. For example, a value at 0.5 on the horizontal axis is the $r^2$ value computed from the poorest half of the datapoints and their corresponding predictions. (b) The horizontal axis specifies a wealth percentile. During training and testing, all data above the wealth percentile is ignored. The vertical axis plots the $r^2$ value between predictions and the true AWI. ``OOC Overall'' corresponds to out-of-country predictions (data is divided into folds by country). Nightlights GBT and Nightlights Ridge operate in the same manner, using gradient boosted trees and ridge regression respectively. ``Pooled'' and ``Block CV Pooled'' correspond to cross validated $r^2$ values. The cross validation is agnostic to the country, so training and testing data may reside in the same country. The ``Block CV Pooled'' model removes any training imagery that overlaps with test imagery.}
\end{figure}

One major finding in \cite{Jean16} was that their convolutional features make much stronger AWI predictions in the poorer segment of the wealth distribution, and we see similar behavior in our models as well. Figure~\ref{fig:poorest_percent} and Figure~\ref{fig:pool_poorest_percent} show the results for the ``VGG-F, 9 Band / ridge'' model. In Figure~\ref{fig:pool_poorest_percent}, we consider the case of training and predicting only on cluster locations that fall below a certain poverty percentile. As in \cite{Jean16} we achieve a significantly higher $r^2$ value than nightlights when training on only the poorer datapoints. Our VGG-F model trained on Landsat 7 imagery surpasses results described in \cite{Jean16} trained on Google Static Maps imagery. However, this only holds when sampling training and test folds in a manner that is agnostic to country borders (marked "Pooled" or "Block CV Pooled" in Figure~\ref{fig:pool_poorest_percent}). We observe that restricting train and test folds to each be exactly the data from a single country results in significantly poorer performance when training on the poorest data (marked "OOC Overall" in Figure~\ref{fig:pool_poorest_percent}).

In Figure~\ref{fig:poorest_percent} we examine leave-one-country-out training with a ridge regression model using image features extracted by the VGG-F 9 Band CNN. We train the model on DHS survey data from 4 of the 5 countries and then have it predict only on datapoints from the left-out country that are below a particular wealth percentile threshold. We see that when the model is applied to countries that it has not seen before, its performance suffers the most in poorer areas.

\section{Conclusion}

Our results show that the current state-of-the-art in satellite-based poverty prediction lends itself to predicting relative wealth within a single country where some ground truth data is available, but may struggle with extrapolating across country borders. Using some combination of nightlights and predictions from the proposed models may yield further improvements. Furthermore, while we only trained models to predict economic livelihood with a single year of Landsat 7 imagery, we could extend our predictions to all of the years that Landsat 7 has been active (since 1999). This opens up the possibility of analyzing changes in local economic levels over time at a much higher granularity than before, especially in developing countries that typically experience long intervals between nationally-representative household surveys.


\begin{table}[t]
\centering
\caption{Results for mean out-of-country predictions. Results are obtained by repeating for each country the process of training on 4 countries and predicting locations in the 5th.  Aggregate Residual $r^2$ indicates the squared correlation between residuals of predictions from nightlights and residuals of predictions from the model, aggregated across all five countries.}
\begin{tabular}{lllll}
  \toprule
  Model & Mean Train $r^2$ & Mean Test $r^2$ & Aggregate Residual $r^2$ \\
  \midrule
  Nightlights / GBT         & 0.63 & 0.66 & 1.0  \\
  VGG-F, RGB / ridge        & 0.71 & 0.64 & 0.69 \\
  VGG-F, 9 Band / ridge     & 0.68 & 0.63 & 0.70 \\
  ResNet-18, 9 Band / ridge & 0.69 & 0.64 & 0.73 \\
  ResNet-34, 9 Band / ridge & 0.70 & 0.65 & 0.74 \\
  \midrule
  Jean et al. \cite{Jean16} & 0.53  & 0.54 & 0.76 \\
  \bottomrule
\end{tabular}
\label{tab:corr}
\end{table}

\clearpage

{\small
\bibliographystyle{ieee}
\bibliography{ms.bib}
}

\end{document}